# Automatic Image Colourizer

Aditya N Parikh
Btech -Entc
Vit-Pune

*Abstract—* In this project we have designed and described a model which colourize a gray-scale image, with no human intervention. We propose a fully automatic process of colouring and re-colouring faded or gray-scale image with vibrant and pragmatic colours. We have used Convolutional Neural Network to hallucinate input images and feed-forwarded by training thousands of images. This approach results in trailblazing results.

## I. INTRODUCTION

Image Colourization is process of reprocessing colour information from the intensity information recovered from the input image. We can notice that even in Gray-scale or black and white images, the semantics of the scene and its surface texture provides ample of cues. For example, the grass is typically green, the sky is typically blue, and the apple is most definitely red. Of course, these kinds of semantic priors do not work for everything. But the actual task is not to recover the originality but rather to produce a plausible result which can fool a human observer. The luminous and texture of an image often gives enough cues for colourization but can also mislead, due to non-variable training data or misleading data, so considering this possible issue, the task to colourize an input seems more achievable.

Texture feature is far more dependable than any other feature detection parameter.

Also, an advantaging property is that we can practically use any training data, i.e., any sample dataset available of any colour, place or object. We just need image's colour channel. We already have lots of training data available and previous works on trained CNN (Convolutional Neural Network) to predict colour, that too, of larger datasets. However, we noticed that most of the previously attempted methods are based on corner-point feature detection and resultant images are highly desaturated. In this approach we have tried to minimize the error between estimated output images and ground truth images.

We also conducted a small survey asking audience to about need for colourization and how satisfied they are with present models available through mobile applications and online software.

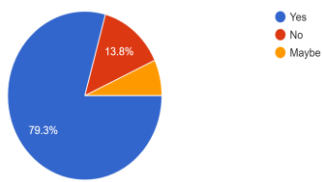

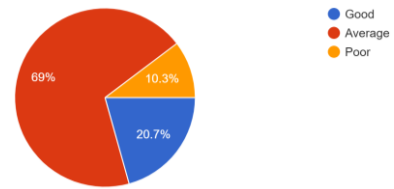

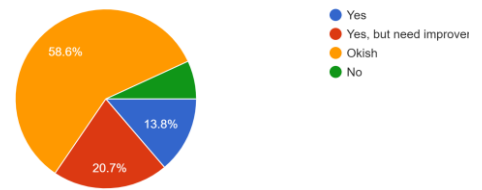

## II. LITERATURE SURVEY

We referred numerous of research paper and techniques for image colourization. Most common way to tackle this problem used is by using OpenCV with CNN and used D-CNN for more accurate results.

Also, Image Colourization using Convolutional Autoencoders. A colour space was taken and implemented with Autoencoders, which is Neural Network's architecture similar to Principal Component Analysis. They are built with two components.

- Encoder
- Decoder

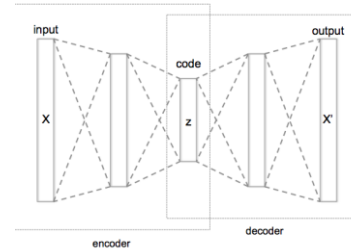

Also, one of the popular approaches is using optimization in Python. YUV/YIQ (Y is the monochromatic luminance channel) colour space is chosen, and algorithm is given as input an intensity volume, considering that similar colour has similar intensity. Pixels with similar intensity are marked with colour scribbles and composing weight matrix to fill colour into neighbouring pixels too.

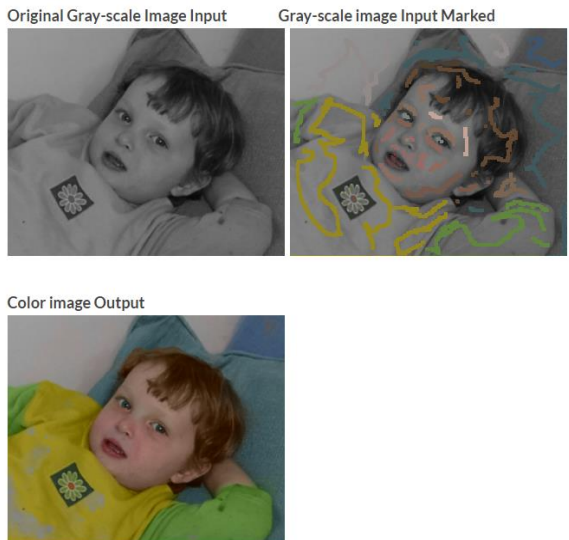

*Fig. An example for optimization techniques by SIGGRAPH paper by Levin et. Al.*

## III. PREVIOUS WORKS AND APPROACHES

Colourizing term was first used for colouring black and white movies into colourful movies, frame by frame, to make it more appealing to the viewers. As mentioned, the process of colourization was done frame by frame manually. Each frame was targeted, manually painted them. After further computer assistance, only painting at least one frame manually and using object tracking techniques propagated the colours throught the movie.

Lot of concurrent work is also going on while using different CNN architecture and loss functions while having same system. Also, the results vary according to the size of dataset and also the overall quality and variety of dataset.

Also repeating convolutional layers again and again also affect the resolution of final outputs.

## IV. APPROACH

We trained CNN to map from given grayscale images, given as input, to a distribution over quantized colour value using architecture given below.

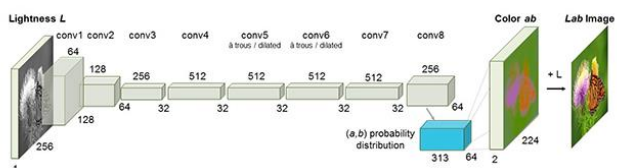

This network architecture was referred from Zhang et al.'s architecture for colourization 2016 ECCV paper.

### A. Objective Function

$X = \mathbb{R}^{h \times w}$ is the input light channel and map it to colour channel $Y = \mathbb{R}^{h \times w \times 2}$, where h and w are dimension of image.

$$L(\hat{Y}, Y) = \frac{1}{2} \Sigma_{h,w} |Y_{h,w} - \hat{Y}_{h,w}|$$

This is equation for Euclidean loss L between predicted and ground truth colours. Where ¥ denotes the predicted outputs.

This loss isn't robust and can vary according to different models. If an object can take on a set of distinct values, the Euclidean loss will be the mean of total set.

### B. Dataset and Colour space

We have extracted training images from ImageNet dataset, images were clustered as we just need colour layer of those images. Diverse images were used so that we have a larger scope while colouring good range of images. For current model we created, we trained 4500 images.

AlexNet architecture directly trained ImageNet dataset to classify it to achieve the highest performance and serves good base for testing. Also, samples of fake-grayscale images were added to that we have gaussian weights and k-means scheme implemented.

The trained model is in release. Caffe model format, which is a deep learning framework developed by Berkeley AI Research (BAIR). This defines a net layer-by-layer in its own model scheme.

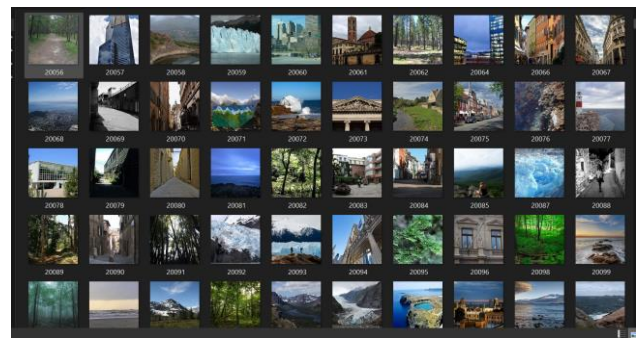

The trained images from RBG colour space to the LAB colour space. Similar to the RGB colour space, the Lab colour has three channels. But unlike the RGB colour space, Lab encodes colour information differently.

- The L channel encodes lightness intensity.
- The a-channel encodes green red.
- And b-channel encodes.

Since the -channel encodes only the intensity, we can use the L-channel as our grayscale input to our system. From that input we can predict a-b channels, then we can predict our final output image.

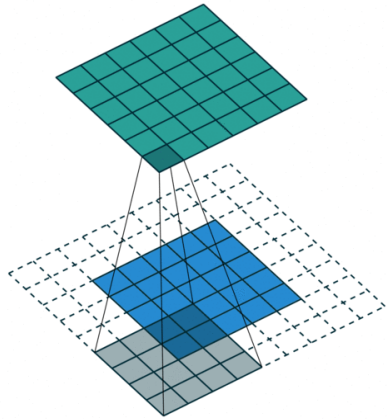

*Fig. Convolutional Neural Network*

### C. Block Diagram – Process

- Convert all training images from RGB colour space to the Lab colour space.
- Use L-channel as the input to network and train to predict ab channels.
- Combine input L-channel with the predicted ab-channel.
- Convert the Lab image back to RGB.

### D. Implementing using GUI

For taking input from a directory and also saving output images, we have designed a simple GUI for our project. We have decided to go with PyGUI to get a simple and minimalistic display to our output, rather than using heavy libraries like Tkinker.

PySimpleGUI is cross-multi-platform GUI to create a nice-looking user interface. We have implemented a directory search box at Left-hand side and a display screen with open-file and save-file toolbox above it at right-hand side.

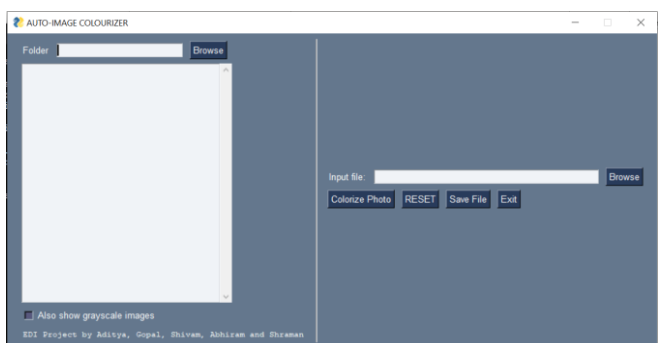

*Fig. PySimpleGUI layout*

## V. RESULTS

### A. Inputs

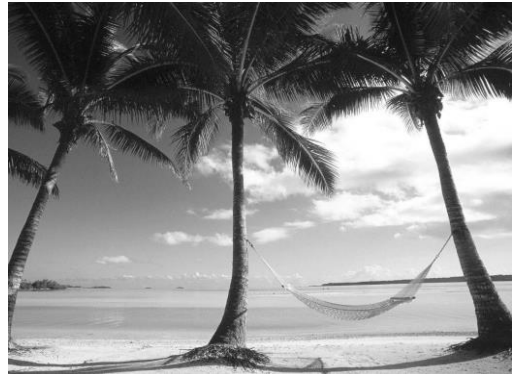

*Fig(i). Input*

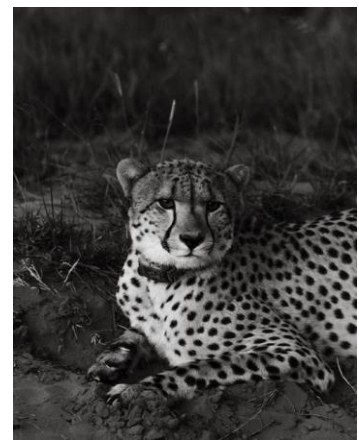

*Fig(ii). Input*

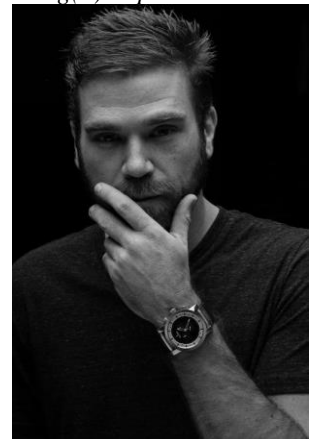

*Fig(iii). Input*

### B. Output

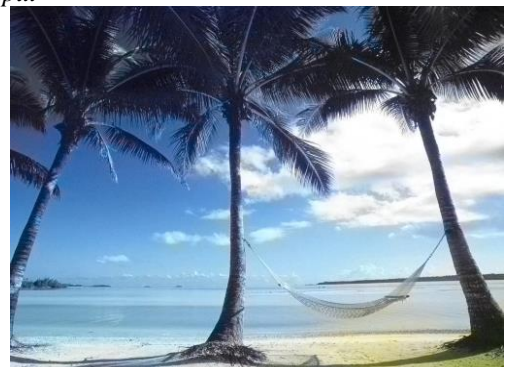

*Fig(i). Output*

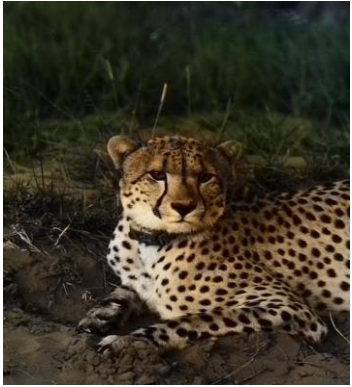

*Fig(ii). Output*

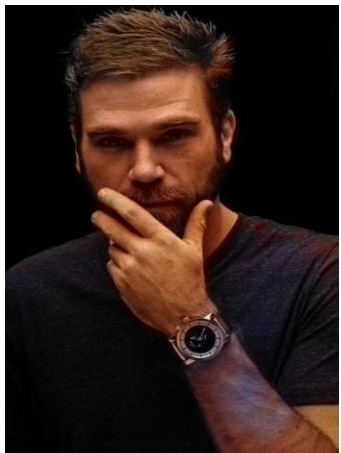

*Fig(iii). Output*

## VI. COMPARISIONS

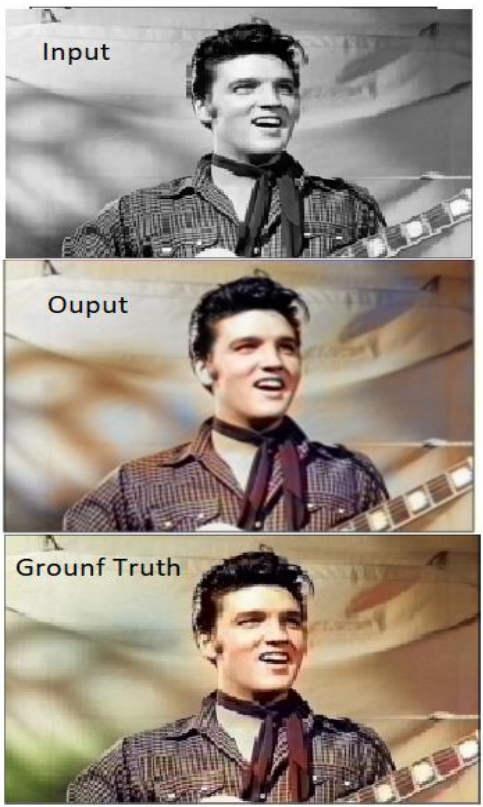

## VII. CONSLUSION

Image colourization being a computer graphic task, has challenged researchers for decades finding close to reality outputs. The model we designed had trailblazing results and can fool a human eye.

This model was developed with texture as important parameter and using CNN to well choose dataset. Also, we found that a Red-colour mask was developed on output image, as a result of invariable sunlight/light source in our dataset. This concludes that, having a large and a quality dataset for training model can lead to even better results.

Image colourization is a field which will be on a verge of improvement as long we will be having better dataset coming and even better if upgrading their optimization techniques.

### REFERENCES

[1] Colorization by Example,R.Irony, D.Cohen-Or, and D.Lischinski, Eurographics symposium on Rendering [2005]
[2] Levin A., Lischinski D.,Weiss Y.: Colorization using optimization. Backtranslations' on Graphics 23, 3 ,689â˜A ,S694 [2004]
[3] Edison. RIUL, Center for Advanced Information Processing, Rutgers University, [2009] http://coewww.rutgers.edu/riul/research/code/EDISON/
[4] Dataset used (all result Images) : LabelMe , Computational Vision Cognition Laboratory (MIT)
[5] T. Leung and J.Malik. Representing and recognizing the visual appearance of materials using three-dimensional textons. International Journal of Computer Vision, 43(1):29-44,[2001]
[6] Li-Jia Li, Hao Su, Eric P. Xing and Li Fei-Fei. Object Bank: A High-Level Image Representation for Scene Classification and Semantic Feature Sparsification. Proceedings of the Neural Information Processing Systems (NIPS), [2010].